# Neutrosophic Ensemble Classification for Uncertainty-Aware Bearing Fault Detection: Evidence from Laboratory and Variable-Speed Industrial Benchmarks

Maikel Leyva-Vazquez 1,2,3*, Dayron Rumbaut Rangel 2, Lorenzo Cevallos-Torres 1, Alexis Matheu Perez 3

1 Universidad de Guayaquil, Guayaquil, Ecuador

2 Universidad Bolivariana del Ecuador, Guayaquil, Ecuador

3 Universidad Bernardo O'Higgins, Santiago, Chile

* Correspondence: maikel.leyvav@ug.edu.ec | ORCID: 0000-0001-7911-5879

Dayron Rumbaut Rangel ORCID: 0009-0001-9087-0979

**Abstract**

Machine learning classifiers for bearing fault detection produce scalar confidence scores that conflate confident errors with genuinely ambiguous predictions, and the conventional truth/falsity pair (F = 1 - T) is algebraically redundant by construction. This paper operationalizes a refined neutrosophic decomposition of a Random Forest + XGBoost + Logistic Regression ensemble into four indicators: T-hat = P1 (top-class evidence), F-hat = P2 (best-competitor evidence, not simply 1 - T-hat), predictive entropy I1-hat (normalized entropy of the full class-probability distribution, which algebraically contains both a within-model and a between-model component and so is not purely aleatoric), and decision disagreement I2-hat (base-learner vote disagreement). We evaluate it on two bearing benchmarks -- CWRU and JNU (600-1000 rpm) -- under a leave-one-condition-out protocol that holds out an entire, previously unseen operating condition per class for testing. On CWRU, after correcting a file-to-class mapping error in which the "Ball" and "Inner" classes had both actually been Inner Race data recorded at two different sampling rates (true Ball-fault files had never been used), the ensemble reaches 100.00% accuracy on three of the four held-out loads, dropping to 92.27% when 0 HP is held out; the three error-free folds leave no errors for the uncertainty indicators to explain, so CWRU serves mainly as a positive control rather than a source of uncertainty-decomposition evidence. On JNU, holding out 1000 rpm, accuracy collapses to 40.64% -- below the 50.03% achieved by always predicting the majority class -- with Logistic Regression (57.91%) generalizing far better than the tree ensembles (30-41%); holding out 600 or 800 rpm instead gives 24.66% and 27.10% respectively, showing the reported fold is JNU's best case, not its average one. T-hat and F-hat remain strongly correlated on JNU (r=-0.910, VIF=5.8) but measurably non-redundant. I1-hat shows a robust, threshold-based association with error (median-split gap +30.4 pp) and a moderate independent linear contribution beyond T-hat/F-hat jointly (partial r=+0.217); a hidden-risk zone -- low-entropy instances where base learners disagree -- shows 63.7% error versus 28.2% when they agree. I2-hat's independent contribution is small and not significant, and in a selective-classification comparison against simple baselines, standalone Logistic Regression confidence outperforms every ensemble-based indicator, including the proposed decomposition. Fusing a time-domain and a frequency-domain model of the same signal and scoring their Jensen-Shannon divergence yields a genuine improvement against that model's own entropy (AURC 0.2868 vs. 0.3649 on the standard split; 0.5415 vs. 0.7338 under a deliberately harder, single-condition reproduction), and, in a CWRU-versus-JNU distributional-shift contrast, is the only indicator in this study that moves in the theoretically expected direction. Separately, on CWRU alone, literature-verified bearing fault frequencies, correctly demodulated via the envelope spectrum, separate most fault classes almost perfectly (99.57%) using only three physically interpretable features, though this does not extend to JNU and cannot demonstrate a selective-classification advantage on CWRU's already error-free fold. Because windows overlap within files and only a few files compose each held-out condition, we report these findings as descriptive associations within this design rather than as claims generalizing to independent replicate conditions.

## 1. Introduction

Rotating machinery is foundational to industrial production. Bearings are among the most failure-prone components: studies consistently attribute 40-50% of electric motor failures to bearing defects [1]. Unscheduled downtime from bearing failure costs manufacturing facilities thousands of dollars per hour, motivating automated predictive maintenance (PdM) systems based on vibration signal analysis.

Machine learning (ML) classifiers trained on time-domain vibration features can identify Normal operation, Ball faults, Inner Race faults, and Outer Race faults with high accuracy under controlled conditions [2][3]. However, a fundamental limitation persists: standard classifiers return a single scalar confidence score that conflates two distinct sources of uncertainty. Confidence magnitude (how far the prediction is from certain) and uncertainty geometry (whether residual probability concentrates on one strong competitor or disperses uniformly across alternatives) provide qualitatively different risk signals for maintenance decision-making.

Neutrosophic logic [4] addresses this by decomposing any proposition into truth (T), falsity (F), and indeterminacy (I). Applied to classification, T and F encode confidence magnitude while I captures uncertainty geometry. Prior work has applied neutrosophic reasoning to classification under uncertainty more broadly [5] and to bearing fault detection specifically [6], but without systematically verifying whether I provides predictive information beyond F -- a prerequisite for justifying the three-component representation.

A second limitation in the bearing fault detection literature is over-reliance on the CWRU benchmark [3], a laboratory dataset [7] whose four standard fault classes are, once correctly identified, easily and near-perfectly separable and so may not reflect real industrial conditions. As we show directly in this paper (Section 4.1), variable rotational speed on the JNU benchmark [8] can reduce classification accuracy substantially -- below the accuracy of a trivial majority-class predictor -- when the test speed is genuinely unseen during training, increasing the practical importance of uncertainty quantification specifically under this kind of distribution shift, rather than on laboratory benchmarks where classes are already easy to separate.

This paper makes five contributions. First, we propose a neutrosophic ensemble pipeline (RF + XGBoost + LR + refined neutrosophic decomposition into T-hat, F-hat, predictive entropy I1-hat, and decision disagreement I2-hat) for bearing fault detection, with I1-hat operationalized as the normalized entropy of the ensemble's averaged class-probability distribution so that it is not an algebraic function of T-hat and F-hat alone, while making explicit that this entropy is not a pure measure of data-inherent (aleatoric) uncertainty, since it algebraically contains a between-model disagreement component as well [16]. Second, we identify and correct a file-to-class mapping error in the CWRU benchmark itself, present in the pipeline before this study and undetected through several earlier correction rounds, in which two of the four nominal classes were actually the same physical fault recorded at different sampling rates; correcting it raises CWRU accuracy to 100.00% on three of its four held-out loads (92.27% on the fourth, Section 4.1, Table 1b) and changes its role in the study from a source of uncertainty-decomposition evidence to a mostly-positive control. Third, we evaluate the pipeline under a leave-one-condition-out protocol, extended here to hold out every available load (CWRU) or speed (JNU) in turn rather than a single one, correcting a window-level data-leakage issue present in earlier versions of this pipeline and showing that the single fold typically reported for JNU is its best case, not its average one. Fourth, we provide statistical evidence, reported as descriptive associations within the tested conditions rather than as claims that generalize to independent replicate conditions, that predictive entropy I1-hat is associated with prediction error beyond T-hat and F-hat jointly on JNU. Fifth, we benchmark the proposed decomposition against simple confidence, margin, and standalone-model baselines in a selective-classification comparison on JNU (CWRU's main 3 HP fold has zero errors to select against once corrected -- Table 1b, Section 4.1), including an oracle joint decision rule over I1-hat and I2-hat jointly, and show that decision disagreement I2-hat is a comparatively weak signal on its own, that jointly thresholding it with I1-hat has no demonstrated operational advantage over I1-hat alone even in the best case, and that standalone Logistic Regression confidence -- not the full ensemble-based decomposition -- gives the best selective-classification performance on JNU. We report this honestly as a boundary condition on the practical, quantitative advantage of the specifically neutrosophic multi-component framing over conventional uncertainty scores, rather than omit it or overstate what has been demonstrated. Sixth, in a newer, less scrutinized part of this study (Section 4.6), we identify two further sources of evidence: fusing two models fit on different feature representations (time versus frequency) of the same signal and scoring their Jensen-Shannon divergence beats that model's own entropy under both the standard split and a deliberately harder single-condition reproduction, and, in a specific CWRU-versus-JNU distributional-shift contrast, is the only indicator in this study that moves in the theoretically expected direction; and, on CWRU alone, literature-verified bearing fault frequencies, correctly demodulated via the envelope spectrum, separate most fault classes almost perfectly using only three physically interpretable features.

## 2. Background and Related Work

Time-domain statistical features for bearing fault detection have been validated extensively. RMS, kurtosis, crest factor, and peak-to-peak amplitude capture fault-induced impulsive patterns at low computational cost [2]. Ensemble methods -- particularly Random Forest and gradient boosting -- achieve near-perfect accuracy on CWRU [3][7], while performance degrades substantially under variable speed conditions due to non-stationarity of fault signatures [8].

Neutrosophic logic [4] extends classical logic with three independent components T, I, F in [0,1]. Kavitha et al. [5] applied neutrosophic classifiers to anomaly detection with threshold optimization. Kumar et al. [6] introduced a neutrosophic cross entropy measure for rolling element bearing fault diagnosis, demonstrating that neutrosophic distance metrics outperform fuzzy-based counterparts on vibration benchmark datasets. However, in classification settings where T = max(P) and F = 1-T, the formulation I_max = 1 - max(P) = F holds algebraically, making one metric redundant [9]. Genuine three-dimensionality requires I to capture distributional properties beyond the point maximum -- such as normalized entropy (I_entropy), margin gap (I_margin), or Gini impurity (I_gini).

Selective prediction frameworks [10] introduce abstention mechanisms but rely on a single uncertainty axis. Deep ensembles [11] estimate uncertainty but impose significant computational overhead for edge deployment. The neutrosophic decomposition proposed here is lightweight (no retraining required), geometrically interpretable, and applicable to any soft-voting ensemble.

## 3. Materials and Methods

### *3.1. Datasets*

CWRU Bearing Dataset [7]. Drive-end accelerometer data under four load conditions (0-3 HP). An earlier version of this pipeline used files 105-108 for "Ball" and 109-112 for "Inner"; per the official CWRU Bearing Data Center catalog these are both Inner Race 0.007" data -- 105-108 at 12 kHz and 109-112 at 48 kHz, the same physical fault at two sampling rates, not two classes -- and the real Ball 0.007" fault (12 kHz) is files 118-121, which had never been used. We correct this: four classes, Normal (files 97-100), Ball fault 0.007" (118-121), Inner Race 0.007" (105-108), Outer Race 0.007" at 6 o'clock (130-133); the 48 kHz duplicate (109-112) is dropped. A second, independent sampling-rate issue affects the Normal class specifically: files 97-100 ("Normal Baseline Data") are recorded at 48 kHz, not 12 kHz like the three fault classes above -- confirmed both by their sample counts (each file's length divided by 48 kHz gives a duration consistent with the dataset's documented recording length, matching the fault files' durations at 12 kHz) and by the published literature's description of a distinct "48k normal-baseline" category for this dataset. Applying the same 1,024-sample window to an unmodified 48 kHz Normal signal would give it one-quarter the physical duration of a fault-class window (21.33 ms vs. 85.33 ms), a sampling-rate confound between Normal and every fault class independent of the Ball/Inner issue above. We resample the four Normal files from 48 kHz to 12 kHz (decimation by 4, with the anti-aliasing filtering scipy.signal.decimate applies) before windowing, so all four classes are windowed from a consistently 12 kHz signal. Sixteen files provide 3,667 windows (1,024 samples, 85.33 ms at 12 kHz, step 512): 824 Normal, 946 Ball, 948 Inner, 949 Outer. (File 99.mat internally bundles an extra drive-end channel belonging to 98.mat, a known artifact of the public CWRU release; we select only each file's own channel.) This is a controlled laboratory benchmark with known fault geometry; motor speed varies only mildly and incidentally with load, from approximately 1797 rpm at 0 HP to approximately 1721 rpm at 3 HP (about 4% overall), so the held-out 3 HP condition (Section 3.3) differs from training conditions primarily in load, with a comparatively small accompanying speed change -- unlike JNU's directly and deliberately varied 600-1000 rpm (about 67% range, described below).

JNU Bearing Dataset [8]. Vibration data from Jiangnan University collected at 50 kHz under three rotational speeds (600, 800, 1000 rpm) using a PCB MA352A60 accelerometer. Fault geometry: 0.3 mm x 0.05 mm artificially induced dents on outer ring, inner ring, and roller. Four classes: Normal (n), Inner race (ib), Outer race (ob), Ball fault (tb). Twelve files provide 17,577 windows (1,024 samples, 20.48 ms at 50 kHz): 8,793 Normal, 2,928 Inner, 2,928 Outer, 2,928 Ball. The variable-speed condition introduces non-stationarity absent in CWRU, producing substantially lower classification accuracy and more errors for statistical analysis. We note that the 1,024-sample window corresponds to a different physical duration on each dataset (Section 5): at 50 kHz it spans only about 0.21-0.34 shaft revolutions over 600-1000 rpm, versus about 2.5 revolutions at CWRU's (resampled) 12 kHz and approximately 1750 rpm.

### *3.2. Feature Extraction*

Twelve time-domain statistical features are extracted per window: mean, standard deviation, RMS, peak value, crest factor (peak/RMS), shape factor (RMS/mean absolute), impulse factor (peak/mean absolute), kurtosis, skewness, peak-to-peak amplitude, variance, and signal energy. Of these, crest factor, shape factor, impulse factor, kurtosis and skewness are scale-invariant by construction; the remaining amplitude-dependent features (mean, RMS, peak, peak-to-peak, variance, energy) are retained for their known discriminative value in the fault-detection literature and are standardized using a StandardScaler fitted on the training partition only, which is applied identically across both datasets despite their different sampling rates.

### *3.3. Train/Test Split: Leave-One-Condition-Out*

An earlier version of this pipeline split windows using a stratified 80/20 random split applied after all sliding windows (1,024 samples, 50% overlap) had already been generated and pooled across files. Because adjacent windows from the same recording share up to 50% of their raw samples, this procedure allowed near-duplicate windows to appear in both the training and test partitions whenever two overlapping windows from the same file were assigned to opposite splits by the random shuffle -- a window-level data-leakage artifact that inflates apparent accuracy and was not detected until a subsequent methodological audit of this pipeline. We correct this with a leave-one-condition-out split: windows are grouped by their source file before any splitting occurs, and an entire operating condition is held out for testing, per class, so that no file -- and therefore no window -- contributes data to both partitions. On CWRU (corrected file mapping, Section 3.1), the main-text held-out condition is the 3 HP load (files 100, 121, 108, 133, one per class); training uses the 0/1/2 HP files. On JNU, the main-text held-out condition is 1000 rpm (files n1000_3_2.csv, ib1000_2.csv, ob1000_2.csv, tb1000_2.csv); training uses 600 and 800 rpm. This yields 2,718 training / 949 test windows on CWRU and 11,718 training / 5,859 test windows on JNU, redistributing the total windows above by condition rather than shuffling them, and constitutes a substantially harder and more realistic generalization test: the classifier must transfer to a load or speed it has never seen, rather than interpolate among conditions already represented in training. Because a single held-out condition could happen to be an easy or hard case rather than a representative one, Section 4.1 additionally reports a multi-condition robustness check that repeats this protocol holding out every available load (CWRU: 0/1/2/3 HP) or speed (JNU: 600/800/1000 rpm) in turn, computing only classification accuracy (not the full statistical battery of Sections 3.5-3.6) for each fold. Section 3.6 and the Limitations (Section 5) discuss the remaining constraints on statistical inference within any single fold.

### *3.4. Ensemble Framework*

The training partition (from the leave-one-condition-out split, Section 3.3) is standardized with a StandardScaler fitted on training data only, then rebalanced with SMOTE [12] applied on the already-scaled training features. An earlier version of this pipeline applied SMOTE to raw, unscaled features before fitting the scaler; because SMOTE synthesizes samples using Euclidean nearest neighbors, and the 12 features (Section 3.2) differ by orders of magnitude in scale, this let high-variance features dominate neighbor selection during oversampling. All results in this manuscript use the corrected scale-then-resample order. Three base learners are then trained independently on the scaled, SMOTE-balanced training partition: (1) Random Forest: 200 trees, Gini criterion; (2) XGBoost [14]: 200 estimators, log-loss evaluation; (3) Logistic Regression: L2 regularization, multinomial softmax. Random Forest itself [13] provides the bagging baseline against which the boosted and linear learners are compared. The ensemble probability P_avg = mean(P_RF, P_XGB, P_LR) preserves the full distributional information required for neutrosophic decomposition.

### *3.5. Refined Neutrosophic Decomposition: Operational Indicators*

We do not interpret the quantities defined below as direct measurements of the philosophical neutrosophic truth-value of a classification. Rather, we treat them as operational indicators: quantities computed from the ensemble's output that are designed to track, but are not equivalent to, the underlying constructs of predictive confidence, opposing evidence, and two distinct sources of indeterminacy. We denote them with a hat (e.g., T-hat) to mark this distinction, following standard construct-vs-operationalization practice in measurement theory.

Under the conventional formulation T = max(P_avg), F = 1 - T, the two components are algebraically redundant (r = -1 by construction), which is why prior neutrosophic ensemble work [9] excluded I_max = F as a degenerate indeterminacy metric. We instead operationalize T-hat and F-hat as evidence for the two leading hypotheses rather than as a value and its complement: T-hat = P_1 (top-class evidence) and F-hat = P_2 (best-competitor evidence), where P_1 >= P_2 are the top two entries of P_avg. We deliberately do not require T-hat + I-hat + F-hat = 1: unlike fuzzy logic, the general neutrosophic framework [4] treats T, I and F as independently valued in [0,1] rather than as a partition of unity. We are careful, however, not to overstate what a correlation r(T-hat, F-hat) different from -1 establishes: it rules out exact algebraic redundancy but does not by itself demonstrate statistical independence, nor rule out a deterministic nonlinear relationship, and P_1, P_2 remain jointly constrained by T-hat >= F-hat and T-hat + F-hat <= 1 regardless of how far r is from -1 (Section 4.2).

An earlier revision of this manuscript defined aleatoric indeterminacy as I1-hat = 1 - (P_1 - P_2), matching the previously validated I_margin operationalization. Subsequent methodological review of that revision showed this choice to be mathematically unsound once F-hat is defined as P_2 rather than as 1 - T-hat: because I1-hat = 1 - (P_1 - P_2) = 1 - T-hat + F-hat is an exact algebraic identity, T-hat + I1-hat - F-hat = 1 always holds, meaning I1-hat carried no information beyond what T-hat and F-hat jointly already encode. We correct this by operationalizing what we now call predictive entropy, I1-hat = -sum_k(P_avg_k * log(P_avg_k)) / log(K), where K is the number of classes. Unlike 1 - (P_1 - P_2), this quantity depends on the probability mass assigned to every class, not only the top two, and for K >= 4 is not expressible as any fixed function of T-hat and F-hat alone: for example, P = (0.50, 0.25, 0.20, 0.05) and Q = (0.50, 0.25, 0.125, 0.125) share the same T-hat = 0.50 and F-hat = 0.25 but have different normalized entropies, 0.840 and 0.875 respectively -- confirming non-redundancy is possible in principle, though not by itself that the extra information is predictively useful, which we test empirically (Section 4.3) rather than assume from the correlation structure alone. We do not call I1-hat "aleatoric" indeterminacy, despite following that convention in an earlier revision: the entropy of an ensemble's averaged distribution decomposes algebraically into a within-model and a between-model term, H(P_avg) = mean_m[H(P_m)] + mean_m[KL(P_m ‖ P_avg)] (a standard identity in the ensemble-uncertainty literature [16]), so I1-hat already contains a disagreement component and cannot be interpreted as purely data-inherent uncertainty. We therefore refer to it plainly as predictive entropy, an operational indicator of overall predictive spread, without claiming it isolates aleatoric from epistemic uncertainty.

Because I1-hat already contains a disagreement term, a second indicator that targets disagreement more directly may add relatively little once I1-hat is controlled for -- an empirical possibility rather than a certainty, and one relevant to interpreting Section 4.5. Following Smarandache's refined neutrosophic logic [15], we nonetheless add a second, operationally distinct indicator alongside I1-hat rather than introducing a fourth axis: I2-hat, which we call decision disagreement rather than "epistemic indeterminacy" for the same reason. We initially operationalized I2-hat as the mean Kullback-Leibler divergence of each base learner's distribution from P_avg, but this measure showed no incremental value beyond T-hat/F-hat/I1-hat and behaved non-monotonically with error rate in early experimentation, because it is sensitive to disagreement on classes that do not affect the final decision. We instead operationalize I2-hat as vote disagreement: the fraction of the three base learners (RF, XGBoost, LR) whose own argmax prediction differs from the ensemble's final predicted class, I2-hat = 1 - (n_agree / 3) in {0, 1/3, 2/3, 1}. This targets disagreement specifically about the decision that matters, rather than distributional distance in general.

*3.6. Statistical Validation Protocol*

Two families of tests are applied to the leave-one-condition-out test set of each benchmark. First, a four-test protocol assesses whether I1-hat provides information beyond T-hat and F-hat jointly: (1) Pearson correlation and Cohen's d (pooled, size-weighted standard deviation) between each indicator and prediction error; (2) band-stratified correlation r(I1-hat, error) within fixed F-hat intervals; (3) a median-split comparison of ERROR RATES (not of I1-hat itself) between instances above versus below the median I1-hat, assessed via a chi-squared test on the resulting 2x2 error/correct contingency table; (4) multivariate partial correlation r(I1-hat, error | T-hat, F-hat), controlling for both simultaneously via linear residualization, to isolate the linear residual contribution -- controlling for F-hat alone would understate what must be tested given that this manuscript's own framework (Section 3.5) treats T-hat and F-hat as two separate, non-redundant indicators. Second, a parallel protocol assesses I2-hat: multivariate partial correlation r(I2-hat, error | T-hat, F-hat, I1-hat), controlling for all three prior indicators simultaneously via linear residualization -- meaningful here specifically because I1-hat (Section 3.5) is no longer a deterministic function of T-hat and F-hat, so the three together span a genuine three-dimensional control set; a three-or-four-level stratification by base-learner agreement (unanimous / two-of-three / one-of-three / zero-of-three); and a hidden-risk analysis comparing error rates among instances with I1-hat at or below the median, split by whether I2-hat indicates any base learner disagreement. All tests use random seed 42 for SMOTE resampling and model initialization; the train/test partition itself is fixed by the leave-one-condition-out design of Section 3.3 rather than by a random seed. For the single historical comparison figure quoted in Section 5 (JNU accuracy under a condition-mixed, non-grouped random split), we re-ran the original pre-redesign pipeline and report the resulting log alongside the other reproducibility artifacts, rather than cite a figure without an accompanying run.

A limitation applies to every correlation and p-value reported in Section 4: the 1,024-sample windows overlap 50% within each source file, so consecutive windows are not independent observations, and each dataset's test partition is drawn from only four files (one per class). We therefore report the statistics below as descriptive summaries of association within this specific test partition, not as inferential claims that would generalize to independently sampled operating conditions; we do not adjust the reported test statistics (e.g., via block-bootstrap or file-level resampling) to compensate for this, and flag it explicitly as a limitation (Section 5) rather than imply a level of statistical power the design does not support. Accuracy and coverage in Section 4.4 are reported descriptively at a fixed, pre-specified grid of abstention thresholds; no threshold is selected or optimized on the test set, and none should be read as validated for deployment. The one exception is the oracle joint decision rule reported in Section 4.4, which by construction selects its best-performing branch using the test set's own errors; we flag it there explicitly as a retrospective, optimistic upper bound, not a deployment-validated policy.

## 4. Results and Analysis

### *4.1. Classification Performance under Leave-One-Condition-Out*

Table 1 summarizes ensemble performance on both benchmarks under the leave-one-condition-out protocol of Section 3.3, using the corrected CWRU file mapping and Normal-class resampling (Section 3.1). On CWRU, holding out the entire 3 HP load, the ensemble achieves 100.00% accuracy (949/949 test windows): once Ball and Inner Race are genuinely distinct classes and Normal is windowed at the same effective sampling rate as the fault classes, the four fault types are easily separable and the classifier generalizes perfectly to this particular unseen load. On JNU, holding out the entire 1000 rpm speed, accuracy is 40.64% (2,381/5,859) -- far below the 82.6% we reproduce under a condition-mixed random split using the original, pre-leave-one-condition-out pipeline (see Section 5) -- which also retains that pipeline's SMOTE-before-scaling order (Section 5), so this comparison is not a clean ablation isolating leakage alone -- and below the 50.03% a classifier achieves by always predicting the majority class (Normal, 2,931/5,859 test windows). Balanced accuracy (50.41%) and macro-F1 (44.46%) are higher than raw accuracy because the ensemble does identify several fault classes reasonably well (Outer recall 0.85) while badly under-recognizing Normal (recall 0.21); raw accuracy alone would understate how much class-specific signal survives the speed shift, and the majority-baseline comparison shows that signal does not yet translate into a net practical improvement over the trivial predictor in overall accuracy terms.

*Table 1. Classification performance under leave-one-condition-out: CWRU (held-out 3 HP, corrected mapping and resampling) vs JNU (held-out 1000 rpm)*

| Class | CWRU Prec. | CWRU Rec. | CWRU F1 | JNU Prec. | JNU Rec. | JNU F1 |
|---|---|---|---|---|---|---|
| Normal | 1.00 | 1.00 | 1.00 | 0.81 | 0.21 | 0.33 |
| Ball | 1.00 | 1.00 | 1.00 | 0.14 | 0.36 | 0.21 |
| Inner | 1.00 | 1.00 | 1.00 | 0.42 | 0.60 | 0.49 |
| Outer | 1.00 | 1.00 | 1.00 | 0.67 | 0.85 | 0.75 |
| Overall acc. | — | — | 100.00% | — | — | 40.64% |
| Balanced acc. | — | — | 100.00% | — | — | 50.41% |
| Macro F1 | — | — | 100.00% | — | — | 44.46% |
| Majority-class | — | — | 25.08%* | — | — | 50.03% |

| | | | | | | | |
|---|---|---|---|---|---|---|---|
| baseline acc. | | | | | | | |

** CWRU test classes are near-balanced (236/236/239/238 of 949); 25.08% = 238/949, from always predicting the TRAINING set's majority class (Outer), which happens to be one of the near-tied classes rather than the test set's own largest class (Inner, 239/949=25.18%). Given the near-balance, the majority-class baseline is uninformative for CWRU and shown only for symmetry with JNU.*

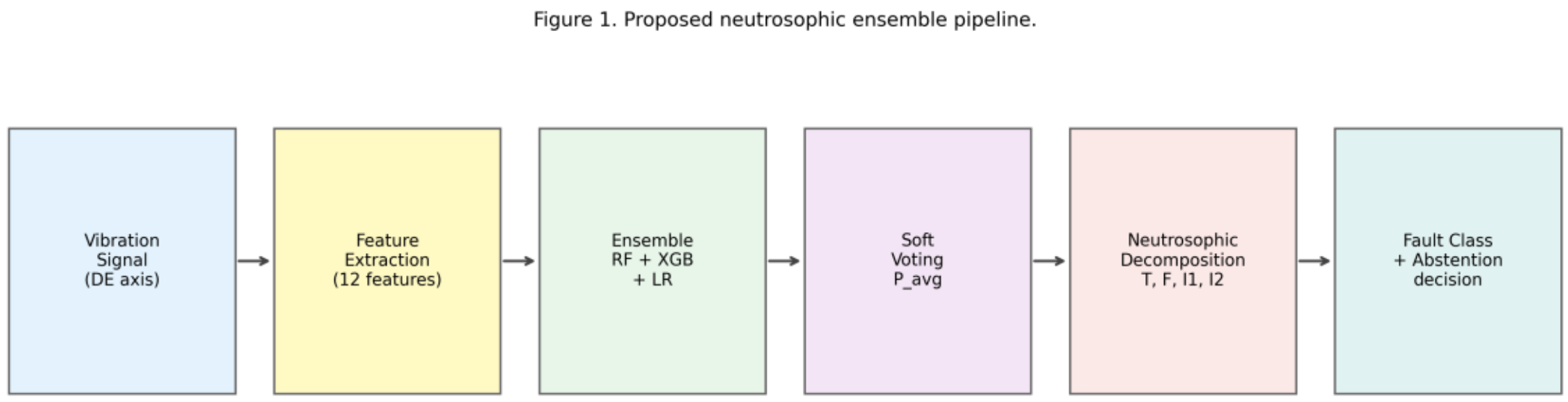

*Figure 1. Proposed neutrosophic ensemble pipeline.*

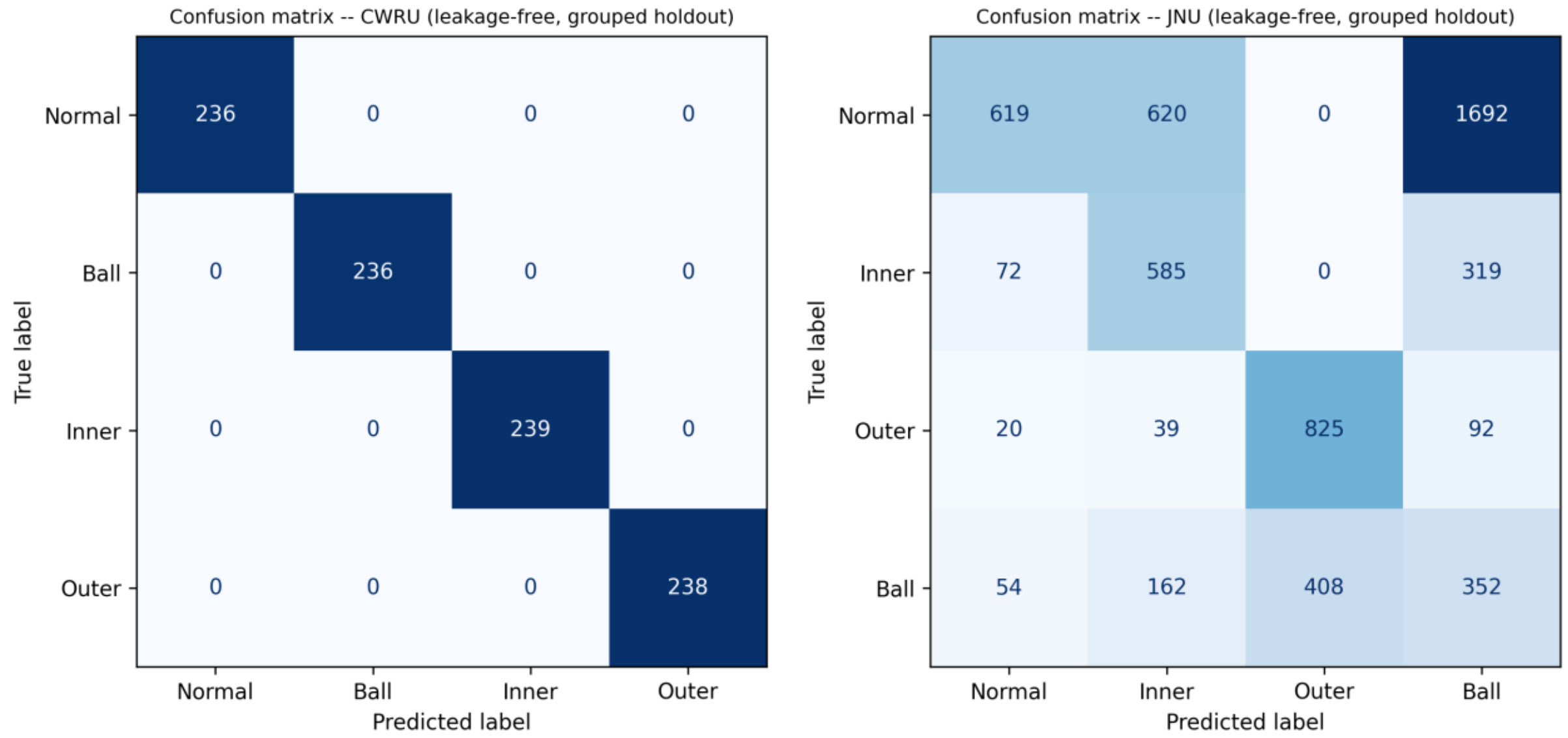

*Figure 2. Confusion matrices on CWRU (left) and JNU (right) test sets, leave-one-condition-out.*

Individual model accuracies on CWRU: RF, XGB and LR all reach 100.00% on the main-text 3 HP fold. On JNU, the picture is qualitatively different: RF 30.07%, XGB 40.72%, LR 57.91%. Logistic Regression generalizes to the unseen speed roughly 17-28 percentage points better than either tree ensemble. Because RF and XGBoost partition feature space using axis-aligned splits learned from the 600/800 rpm training distribution, they likely overfit to speed-specific feature thresholds that do not transfer to 1000 rpm, whereas the linear decision boundary of Logistic Regression extrapolates more gracefully. The soft-voted ensemble (40.64%) sits close to XGBoost and well below LR, meaning equal-weight averaging is not the best strategy under this kind of distribution shift -- a limitation we return to in Section 5, and one that motivates the baseline comparison of Section 4.4.

Because a single held-out condition could be an unrepresentative best or worst case rather than a typical one, we repeated this protocol holding out every available load (CWRU) or speed (JNU) in turn (Table 1b), reporting only ensemble accuracy, balanced accuracy, macro-F1 and individual-model accuracy per fold (the full statistical battery of Sections 3.5-3.6 is applied only to the main-text fold, for the reasons discussed in Section 3.6 and Section 5). On CWRU, accuracy is 100.00% on three of the four folds, but drops to 92.27% when the 0 HP load is held out instead -- a real, non-negligible generalization gap driven mainly by Logistic Regression (68.84% on that fold, dragging down an otherwise strong RF/XGB pair), showing that the 3 HP result, while representative of most folds, is not universal: CWRU is not trivially perfect under every possible held-out load. On JNU, accuracy varies substantially by held-out speed: 24.66% (600 rpm), 27.10% (800 rpm), and 40.64% (1000 rpm) -- meaning the 1000 rpm fold reported as the main result throughout this paper is JNU's best case, not its average one, and the true difficulty of generalizing to an arbitrary unseen speed on this benchmark is, if anything, understated by our headline number.

*Table 1b. Multi-condition robustness check: ensemble accuracy holding out each available load (CWRU) or speed (JNU) in turn*

| **Held-out condition** | **N test** | **Ensemble acc.** | **Bal. acc.** | **Macro F1** | **RF / XGB / LR** |
|---|---|---|---|---|---|
| CWRU: 0 HP | 828 | 92.27% | 93.28% | 90.78% | 100.00% / 98.91% / 68.84% |
| CWRU: 1 HP | 946 | 100.00% | 100.00% | 100.00% | 100.00% / 100.00% / 99.89% |
| CWRU: 2 HP | 944 | 100.00% | 100.00% | 100.00% | 100.00% / 100.00% / 100.00% |
| CWRU: 3 HP (main text) | 949 | 100.00% | 100.00% | 100.00% | 100.00% / 100.00% / 100.00% |

| **Held-out condition** | **N test** | **Ensemble acc.** | **Bal. acc.** | **Macro F1** | **RF / XGB / LR** |
|---|---|---|---|---|---|
| JNU: 600 rpm | 5,859 | 24.66% | 37.01% | 25.13% | 22.29% / 23.50% / 30.28% |
| JNU: 800 rpm | 5,859 | 27.10% | 40.49% | 30.48% | 25.38% / 26.63% / 70.92% |
| JNU: 1000 rpm (main text) | 5,859 | 40.64% | 50.41% | 44.46% | 30.07% / 40.72% / 57.91% |

### *4.2. Non-Redundancy of T-hat and F-hat*

Table 2 reports the correlation and effect size (Cohen's d, using pooled, size-weighted standard deviation) of each operational indicator against prediction error, and the T-hat/F-hat correlation, on JNU (CWRU's main 3 HP fold has zero errors to analyze; its 0 HP fold does have errors but is not analyzed at this granularity -- Table 1b, Section 4.1). r(T-hat, F-hat) = -0.910: strongly negative, since F-hat is mechanically bounded by the probability mass left after T-hat, but no longer the exact r=-1.000 that the classical F=1-T formulation guarantees by construction, and consistent with F-hat tracking real evidence for a specific competing class. As Section 3.5 notes, this departure from -1 specifically rules out the classical F=1-T identity, not every possible algebraic or nonlinear dependence, and a strong linear relationship remains: the variance inflation factor between T-hat and F-hat is VIF = 1/(1-r^2) = 5.8, indicating real but moderate collinearity (a conventional rule of thumb flags VIF > 10 as severe), which warrants caution when interpreting the sign of any partial association involving both variables (Section 4.3). All four indicators separate errors from correct predictions with small-to-moderate effect sizes (|d| approx. 0.6-1.0), reflecting that JNU errors are pervasive rather than confined to a narrow high-uncertainty tail. On CWRU, the indicators themselves remain well-defined, but their correlation with error is undefined: with zero misclassifications on the main-text 3 HP fold (Section 4.1) there is no error variance for any indicator to explain, so r(T-hat, F-hat) = -0.990 there describes the indicators' own relationship but has no accompanying error-correlation to report.

*Table 2. Correlation of each operational indicator with prediction error (Cohen's d, Pearson r) on JNU; T-hat/F-hat correlation on both*

| **Indicator** | **JNU d** | **JNU r** |
|---|---|---|
| T-hat | -0.873 | -0.394 |
| F-hat | +0.585 | +0.276 |
| I1-hat (predictive entropy) | +1.006 | +0.443 |
| I2-hat (decision disagreement) | +0.642 | +0.301 |
| r(T-hat, F-hat): JNU -0.910 (VIF=5.8) \| CWRU -0.990 (indicators defined; error-correlation undefined, zero errors) | | |

*4.3. Statistical Analysis of Predictive Entropy (I1-hat) beyond T-hat and F-hat*

This analysis applies to JNU only: with zero CWRU errors (Section 4.1), there is no error variance for I1-hat, or any other indicator, to explain there. Table 3 summarizes the four-test validation protocol for I1-hat (predictive entropy) on JNU. The raw correlation of I1-hat with error is positive (Table 2). The band-stratified analysis, which asks whether I1-hat still discriminates error within narrow bands of F-hat, is positive in every band and significant in four of five (not significant in [0.40,0.60), r=+0.046, p=0.483, N=233; strongest in [0.10,0.20), r=+0.638, p=6.4e-72, N=617); with five bands tested and no multiple-comparisons correction applied (Section 5), the weaker bands should be read cautiously. The median-split test -- comparing error rates, via a chi-squared test, between instances above versus below the median I1-hat -- shows a clear practical gap: +30.4 pp (74.6% vs. 44.2% error, p=8.8e-124). The multivariate partial correlation r(I1-hat, error | T-hat, F-hat), which controls for T-hat and F-hat jointly rather than F-hat alone (Section 3.6) -- an important distinction given their VIF=5.8 collinearity (Section 4.2) -- isolates a moderate, positive independent contribution (r=+0.217, p=3.9e-63), consistent with I1-hat adding information beyond T-hat and F-hat under genuine distribution shift, though substantially smaller than the raw correlation alone would suggest. As noted in Section 3.6, these results describe association within this specific test partition and should not be read as inference to independently sampled operating conditions.

*Table 3. Four-test validation protocol for I1-hat beyond T-hat and F-hat: JNU (leave-one-condition-out)*

| Test | JNU |
|---|---|
| Band-stratified r(I1-hat,error\|F-hat band) | Positive in 5/5 bands, significant in 4/5; [0.40,0.60) n.s. (p=0.483, N=233); strongest [0.10,0.20) r=+0.638, p=6.4e-72 |
| Median-split error-rate gap (pp) | +30.4 pp (high=74.6%, low=44.2%; chi2 p=8.8e-124) |
| Multivariate partial r(I1-hat, error \| T-hat, F-hat) | +0.217 (p=3.94e-63) -- moderate, positive independent contribution beyond T-hat and F-hat |

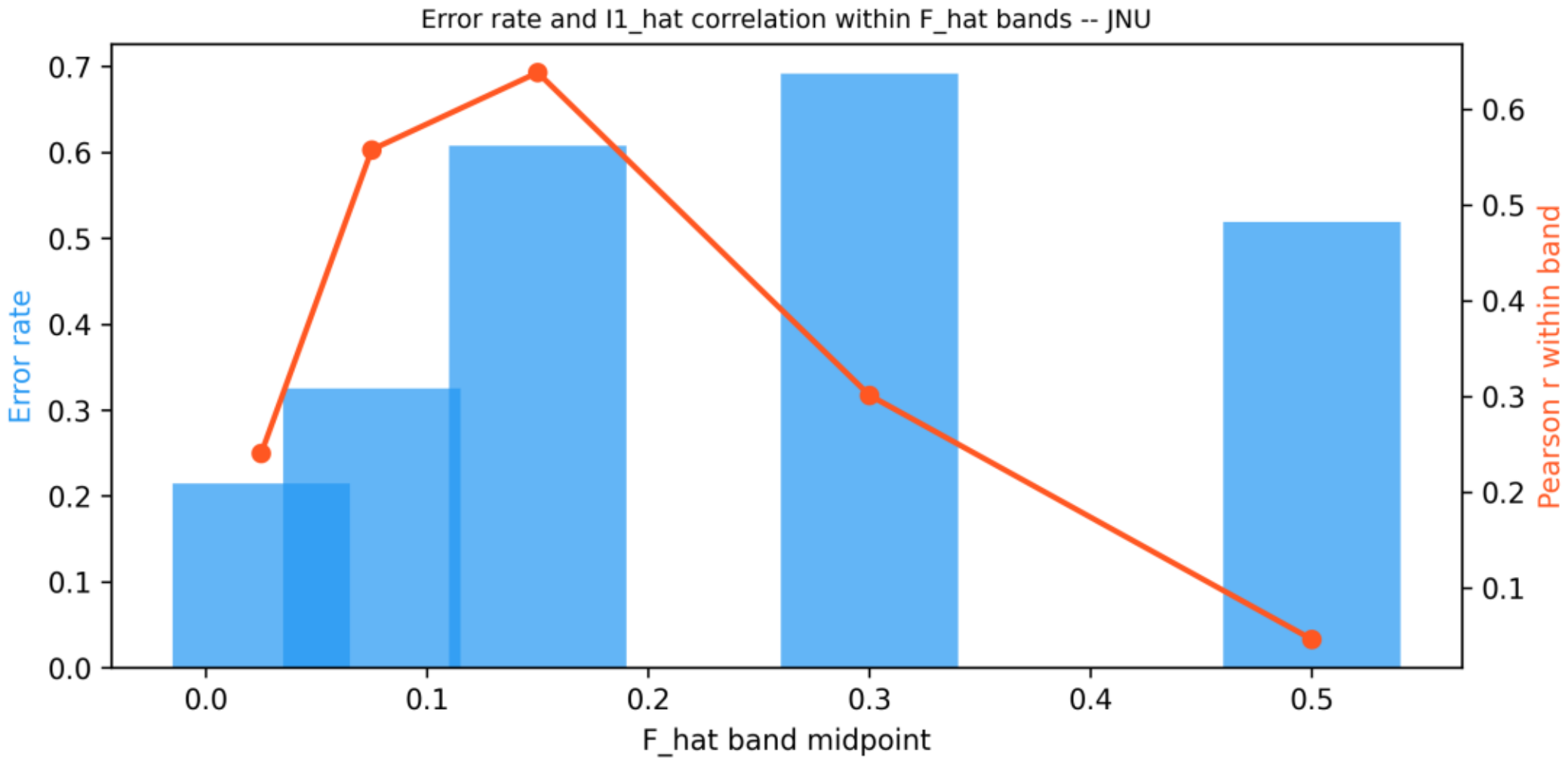

*Figure 3. Error rate and I1-hat correlation within F-hat bands (JNU).*

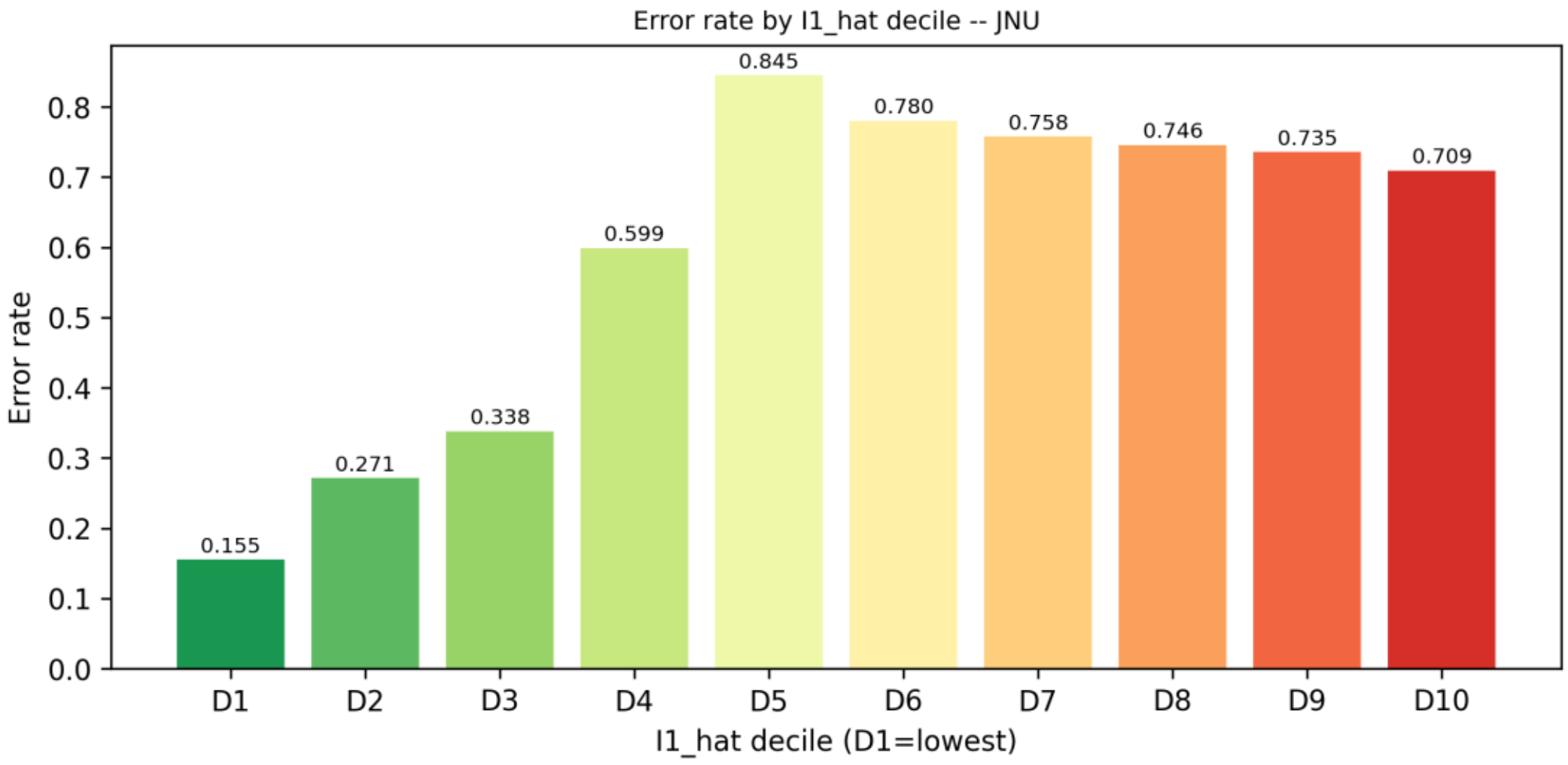

*Figure 4. Error rate by I1-hat decile (JNU).*

The decile analysis (Figure 4) shows a clearly non-monotonic pattern on JNU: error rate rises sharply from D1 (15.5%) to a peak at D5 (84.5%), then decreases and oscillates across the highest deciles (D6-D10 ranging 71-78%). I1-hat separates the lowest-risk decile from the rest clearly, but does not cleanly separate the highest-risk instances from the merely high-risk majority -- a property we report accurately rather than characterize as a steady increasing trend.

*4.4. Selective Risk-Coverage Analysis and Baseline Comparison*

This analysis applies to JNU only (CWRU's main 3 HP fold has no errors to trade off against coverage, Section 4.1). Figure 5 shows the standard selective-classification risk-coverage curve (Geifman and El-Yaniv [10]): risk (error rate among accepted instances) as a function of coverage, for five candidate selectors, each accepting instances in order of increasing uncertainty. An earlier version of this analysis plotted accuracy and coverage separately against a threshold tau and reported a utility-maximizing tau* obtained by maximizing accuracy x coverage on the test set itself; we removed this because the product is algebraically degenerate -- since the accepted set only grows as tau increases, accuracy x coverage reduces to (correct-and-accepted count)/N, which is monotonically non-decreasing in tau by construction, so its maximum always sits near 100% coverage regardless of the data and cannot represent a genuine trade-off. Figure 5 plots the conventional risk-vs-coverage form instead, with no threshold optimization involved.

To assess what the proposed decomposition adds beyond simpler alternatives, we compare I1-hat (predictive entropy) and I2-hat (decision disagreement) against three baselines evaluated on the same ensemble predictions: max confidence (1 - T-hat), margin (1 - (T-hat - F-hat)), and the standalone Logistic Regression model's own confidence (using LR's predictions and errors directly, not the ensemble's). Because I2-hat takes only four discrete values, a plain sort would break ties among thousands of instances in an arbitrary, unstated order; AURC and accuracy figures below use each tied group's expected risk under uniform-random tie-breaking instead (baseline_comparison_jnu.py). Table 3b reports the area under the risk-coverage curve (AURC; lower is better) over the full coverage range, and accuracy at a fixed 50% coverage reference point. I1-hat (AURC=0.3915) is the best-performing selector among those built from the RF+XGB+LR ensemble's own output, modestly ahead of max confidence (0.4050) and margin (0.4220), and ahead of I2-hat (0.4452, the worst of the five). However, standalone Logistic Regression's own confidence (AURC=0.3773) achieves a lower AURC than every ensemble-based selector, including I1-hat -- consistent with LR itself generalizing to the unseen speed far better than the ensemble (Section 4.1); we report this as a lower AURC, not as evidence of better probability calibration, which we did not separately evaluate. This is a boundary condition for the practical value of the proposed decomposition: it improves on naive ensemble confidence, but the added complexity of a three-model ensemble plus a four-indicator decomposition does not clearly beat simply deploying the single best-generalizing base model and using its own confidence for selective prediction under this distribution shift.

A further question is whether treating I1-hat and I2-hat as two SEPARATE axes -- rather than one score -- can outperform I1-hat alone through a genuinely joint decision rule, rather than the simple linear sum tested above. We tested this directly: for each of I2-hat's four discrete values tau2, we computed the exact tie-corrected risk-coverage curve of the AND-rule "accept if I1-hat <= tau1 and I2-hat <= tau2" at every one of that subset's own data points (the same expected-risk-under-random-tie-breaking method used for every other curve in Table 3b, not a coarser quantile grid), then took the pointwise minimum across the four curves as the joint rule's own frontier -- an oracle upper bound on what any AND-rule over these two axes could achieve on this test set, computed on an identical footing to the single-score baselines. This gives AURC=0.391497, against 0.391543 for I1-hat alone: a difference of 0.000046 (about 0.01% relative; both figures and the coverage count below are printed directly by joint_decision_rule_jnu.py, not computed separately), with the joint frontier strictly better than I1-hat alone at only 514 of 5,859 coverage levels and never worse (an envelope over a superset of rules cannot underperform any single member of that set, I1-hat alone included). This oracle frontier selects its best-performing branch using the same test set's own errors, making it a retrospective, optimistic upper bound rather than a policy validated for deployment on unseen data -- unlike every other threshold in this paper, which is fixed before looking at outcomes. We did not run a significance test on this gap and do not claim it is distinguishable from zero; we report it as: the aggregate AURC improvement is very small (0.000046), and we have not evaluated operating costs or a minimum difference we would consider meaningful, so its operational relevance remains undemonstrated rather than ruled out. An earlier version of this analysis, using a coarser threshold grid for the joint search than the exact method used for the single-score curves, had reported an apparent (and, on inspection, purely numerical) AURC of 0.3923, i.e. slightly worse than I1-hat alone -- a comparison invalidated by the grid mismatch itself rather than by any property of the joint rule, which we correct here. Taken together with the ensemble-vs-LR-alone finding above, this directly bears on what the neutrosophic framing contributes here: the four indicators are computed with entirely conventional tools (top-two ensemble probabilities, Shannon entropy, vote disagreement), and neither a linear combination nor an oracle-optimal joint threshold rule over them has a demonstrated operational advantage over the single best conventional score (entropy) on this benchmark. What the framing offers, on the present evidence, is an organizational and interpretive structure -- separating confidence magnitude, competing-class evidence, and two forms of indeterminacy for diagnostic purposes (Section 4.5's hidden-risk zone remains informative as a qualitative flag even where it does not meaningfully improve AURC) -- rather than a demonstrated quantitative advantage over conventional uncertainty quantification, a distinction we did not draw sharply enough in earlier framing of this work.

*Table 3b. Selective-classification baseline comparison on JNU: area under the risk-coverage curve (AURC, lower is better) and accuracy at 50% coverage*

| **Selector** | **AURC** | **Accuracy @ 50% coverage** |
|---|---|---|
| Max confidence (T-hat) | 0.4050 | 53.40% |
| Margin (T-hat - F-hat) | 0.4220 | 50.32% |
| I1-hat: predictive entropy | 0.391543 | 55.82% |
| I2-hat: decision disagreement | 0.4452 | 52.09% |
| I1-hat + I2-hat (linear combination) | 0.3968 | 55.58% |
| I1-hat AND I2-hat (exact oracle joint rule, best case) | 0.391497 | 55.82% |

Logistic Regression alone (not ensemble): AURC=0.3773, accuracy @ 50% coverage=57.49% -- the best result in this comparison, using only a single base model's own confidence.

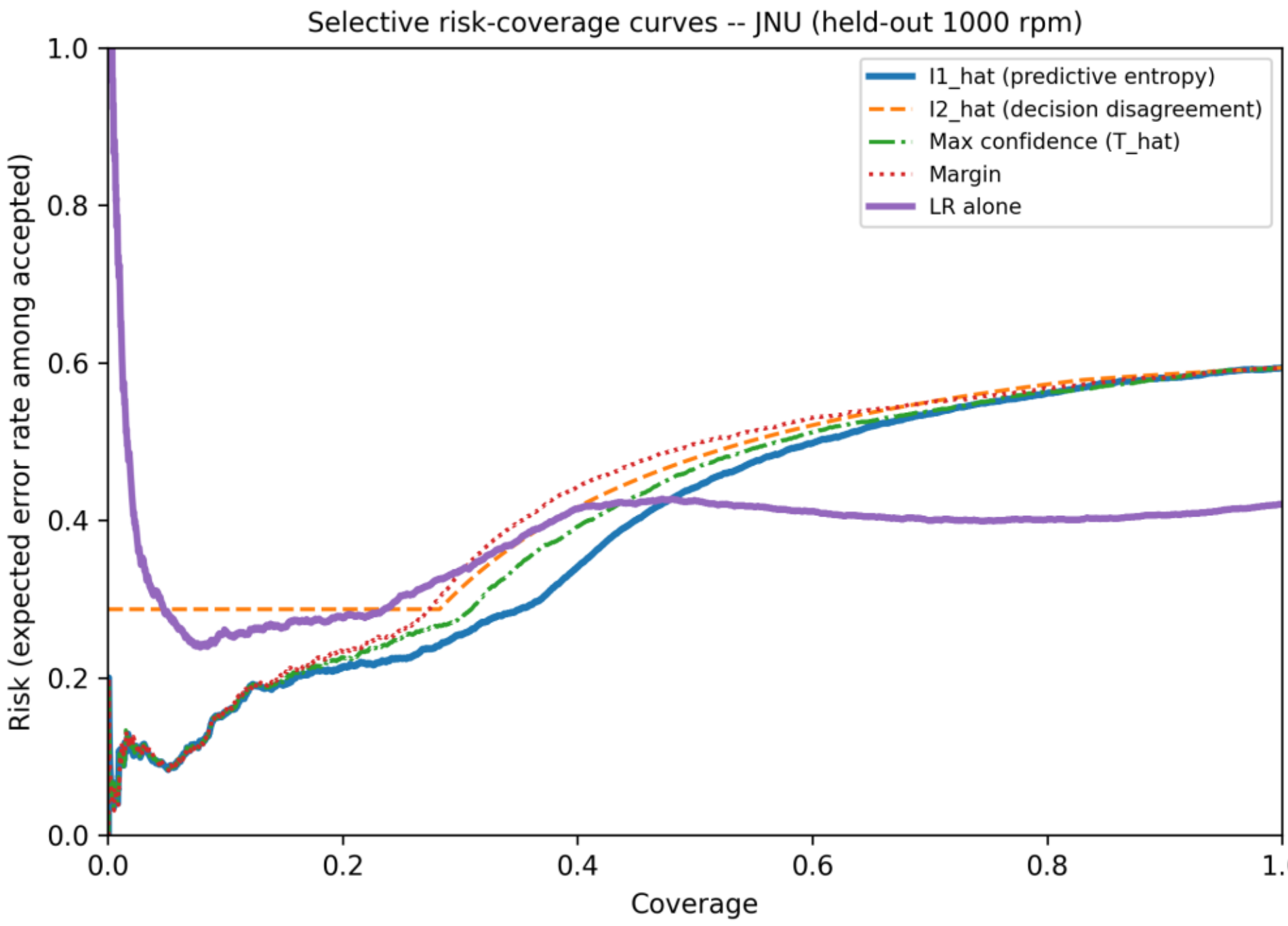

*Figure 5. Selective risk-coverage curves on JNU: I1-hat, I2-hat, max confidence, margin, and standalone Logistic Regression.*

### *4.5. Decision Disagreement (I2-hat)*

This analysis applies to JNU only; CWRU's main 3 HP fold has zero errors (Section 4.1), so partial correlations involving error are mathematically undefined there (no error variance for any indicator to explain). Any non-empty CWRU subgroup's error rate is simply zero, not undefined; only a genuinely empty subgroup (zero instances) would give an undefined rate. We evaluated I2-hat (vote disagreement among RF, XGBoost, LR) using the multivariate partial correlation protocol of Section 3.6, controlling for a genuinely three-dimensional [T-hat, F-hat, I1-hat] set (Section 3.5). Partial r(I2-hat, error | T-hat, F-hat, I1-hat) = +0.017 (p=0.20, a nominal value computed under an independence assumption the 50%-overlapping windows do not fully satisfy, Section 3.6): correctly signed but too small relative to its uncertainty to treat as a reliable independent contribution, given I1-hat already contains a between-model disagreement component (Section 3.5) that captures much of what I2-hat could add linearly. Agreement is far from unanimous under genuine distribution shift (unanimous 1,655/5,859 = 28.2%, error rate 28.7%; two-of-three 3,203/5,859 = 54.7%, error rate 72.9%; one-of-three 976, error rate 66.9%; a zero-of-three group of 25 instances, error rate 64.0%) -- disagreement is the norm rather than the exception here, yet this does not translate into a clearly nonzero independent linear contribution once T-hat, F-hat and I1-hat are accounted for.

*Table 4. I2-hat (decision disagreement) validation on JNU under leave-one-condition-out*

| Indicator | JNU |
|---|---|
| Partial r(I2-hat, error \| T,F,I1) | +0.017, p=0.20 (n.s.) |
| Error rate: unanimous (agree=3/3) | 28.7% (N=1,655) |
| Error rate: exactly 2/3 agree | 72.9% (N=3,203) |
| Error rate: exactly 1/3 agrees | 66.9% (N=976) |
| Hidden-risk zone (low I1-hat, disagreement) | 28.2% (agree, N=1,612) vs. 63.7% (disagree, N=1,318) |

The hidden-risk analysis compares error rates among low-I1-hat instances, split by whether base learners agree. Among the 2,930 JNU test instances with I1-hat at or below the median, the 1,612 where base learners agree show 28.2% error, while the 1,318 where they disagree show 63.7% error -- a substantial gap, directionally consistent with I2-hat's raw correlation (Table 2) even though its partial contribution beyond T-hat/F-hat/I1-hat is not significant (Table 4). On CWRU this comparison cannot be run at all on the main 3 HP fold: among low-entropy instances, the disagreeing group is empty (N=0, undefined rate), and the agreeing group (N=475) has a well-defined error rate of exactly 0%: the comparison fails because there is no disagreeing group to compare against, not because both rates are undefined. We regard the JNU finding as informative in its own right, not as one half of a cross-dataset contrast, since CWRU under the corrected mapping no longer provides a comparison case for this analysis.

Taken together, I2-hat's independent linear contribution beyond T-hat, F-hat and I1-hat on JNU is small and not significant, consistent with the baseline comparison of Section 4.4 in which I2-hat was the weakest selective-classification indicator tested. This is plausibly explained by Section 3.5's algebraic point: I1-hat already contains a between-model disagreement term, so a separate vote-disagreement indicator may have limited additional linear information to contribute once I1-hat is included, though the practically large hidden-risk gap (28.2% vs. 63.7%) shows I2-hat still carries some signal that a purely linear partial-correlation test can understate. We do not have a CWRU comparison to assess whether this pattern is dataset-general, which is itself a limitation of having only one benchmark on which errors remain to analyze.

For completeness, we also note that an initial operationalization attempt for I2-hat, using mean KL-divergence of each base learner from P_avg (a natural distributional analogue to epistemic uncertainty in deep ensembles [11]), showed the wrong-signed partial correlation and a non-monotonic decile pattern in early experimentation preceding the leave-one-condition-out redesign. We do not report exact figures for this attempt here, since we could not reproduce a single, unambiguous numeric result for it from the scripts retained from that stage of the project; we record the qualitative outcome -- KL-divergence did not hold up as an operationalization and was abandoned in favor of vote disagreement -- for transparency about what was tried and did not work, without asserting decimal values we cannot independently re-verify.

*4.6. Two Further Findings: Heterogeneous-Domain Fusion and Literature Bearing-Frequency Validation*

Sections 4.4-4.5 establish an honest negative result: on the evidence gathered so far, neither a linear combination nor an oracle joint rule over I1-hat and I2-hat has a demonstrated operational advantage over predictive entropy alone. We report two further, newer, less scrutinized tests aimed directly at that open question, neither of which has been through this manuscript's adversarial review process.

First, heterogeneous-domain fusion. Every T-hat/F-hat/I-hat construction elsewhere in this paper derives from a single feature representation -- the same 12 time-domain statistics -- read by three base learners that see identical numeric inputs. We instead trained two separate Logistic Regression models on two different feature representations of the same signal: Model A on the original 12 time-domain features, and Model B on 11 frequency-domain features (spectral centroid, bandwidth, entropy, kurtosis, dominant frequency and its relative energy, and four frequency-quartile band energies); both are deterministic transformations of the same sensor channel and window, so we describe them as heterogeneous feature representations, not independent physical evidence sources. Scoring the Jensen-Shannon divergence I between Model A's and Model B's predicted distributions against Model A's own errors, I reaches AURC=0.2868 on the standard split, better than Model A's own entropy (0.3649), with the strongest partial correlation with error in this study (r=0.3088, p=1.23e-129, controlling for Model A's own top-two probabilities; a separately defined heterogeneous alternate-class-evidence score reaches an even better AURC of 0.2032 by exploiting Model B's higher raw accuracy, so we do not claim I is the strongest possible construction). Repeating the test with both models fit on 600 rpm alone and evaluated on 1000 rpm -- a harder single-condition fit, not a fully independent validation, since no decision was selected on an intermediate partition -- I again beats Model A's own entropy (AURC=0.5415 vs. 0.7338; partial r=0.2085, p=1.58e-58). Feeding genuine out-of-distribution CWRU signals through the JNU-fit models, I rises sharply (+0.3285) and A/B agreement collapses from 52.55% to 2.89%, a large, correctly-signed mean-shift contrast (not a calibrated OOD detector); the original entropy-based indicators, evaluated under this same genuine shift, instead decrease -- a confident-extrapolation failure mode that this construction does not share (heterogeneous_fusion_jnu.py, validated_heterogeneous_fusion_jnu.py).

Second, a physically-motivated check on CWRU alone, since JNU's bearing geometry is not confirmed in the primary source material available to us. CWRU uses a documented SKF 6205-2RS JEM bearing, for which the outer-race, inner-race and ball-spin defect-frequency multipliers of shaft speed (BPFO=3.5848, BPFI=5.4152, BSF=2.357) are published. Computing narrowband energy at these literature frequencies directly from each window's raw FFT failed, and in the wrong direction (highest energy in the Normal class; 67.76% accuracy from a 3-feature classifier, far below the standard 12-feature ensemble's 100% on this fold) -- a known pitfall, since bearing-defect energy appears as amplitude-modulation sidebands in the envelope spectrum, not as narrowband energy in the raw spectrum. A four-way ablation isolated this transform effect from a concurrent window-length change: the envelope spectrum, at the SAME window length that failed with the raw spectrum, reaches 97.47% (vs. 67.76%), with a longer window adding a further, smaller improvement to 99.57% (232 of 233 test windows correct). The expected dominant-band pattern holds clearly for Outer and Inner; Ball is discriminated well by the 3-feature classifier overall but does not show a clean one-band correspondence. Because CWRU's standard ensemble is already error-free on this fold, this result cannot demonstrate a selective-classification advantage; we report it as physically-motivated confirmation, obtained with only three interpretable features, that most of these classes correspond to their literature-predicted bearing fault mechanisms (literature_bpfo_bpfi_cwru.py, envelope_bpfo_bpfi_cwru.py).

## 5. Discussion

Developing this pipeline required correcting several distinct classes of methodological problem, none of which were visible from the headline accuracy numbers alone -- itself a methodological point we return to below: (1) a window-level data-leakage issue (overlapping windows from the same file could appear in both train and test under a condition-mixed random split); (2) a definitional flaw in the aleatoric indicator itself (I1-hat = 1-(P1-P2) was an exact algebraic function of T-hat and F-hat, and so could not carry independent information by construction, regardless of the train/test split used), together with several statistical testing errors (a tautological median-split test, an unweighted Cohen's d, incorrect degrees of freedom, and a partial correlation that controlled for F-hat alone rather than T-hat and F-hat jointly); (3) a resampling-order artifact (SMOTE applied before feature scaling) and an algebraically degenerate abstention "utility" metric; (4) a CWRU file-to-class mapping error present from before this study in which the nominal "Ball" and "Inner" classes were both Inner Race data at different sampling rates, and the real Ball-fault files had never been used (Section 3.1); and (5) a second, independent CWRU sampling-rate issue in which the Normal class (recorded at 48 kHz) was windowed identically to the fault classes (12 kHz) without resampling, giving Normal windows one-quarter the physical duration of a fault window. We discuss the substantive findings of the fully corrected pipeline below.

The most consequential single result of this study is that CWRU, once its classes are correctly defined and consistently sampled, is not a source of uncertainty-decomposition evidence at all: the ensemble reaches 100.00% accuracy on three of its four held-out loads (Table 1b), consistent with the literature's general characterization of CWRU as a near-perfectly separable laboratory benchmark once genuinely distinct fault classes are used [3]. This is not unconditional, however: holding out 0 HP instead gives 92.27%, driven mainly by Logistic Regression (68.84% on that fold), so CWRU is not trivially perfect for every possible held-out load, and a claim of perfect generalization should be scoped to the specific folds where it holds. This reframes CWRU's role in the study from a second uncertainty-decomposition testbed to a mostly-positive control confirming the pipeline and features work correctly when classes are well-defined, consistently sampled, and conditions are similar (CWRU's motor speed varies only about 4% with load, Section 3.1). All of this paper's substantive uncertainty-decomposition findings therefore come from JNU alone (using the zero-error 3 HP fold as CWRU's main-text condition), which is also why we no longer report paired CWRU/JNU comparisons in Sections 4.2-4.5.

On JNU, holding out 1000 rpm gives 40.64% accuracy -- far below the 82.6% we reproduce, and report with its own log for reproducibility (Section 3.6), by re-running the original condition-mixed random-split pipeline that predates the leave-one-condition-out redesign, and even below the 50.03% a trivial majority-class predictor achieves (Section 4.1). The multi-condition check (Table 1b) shows this is JNU's best-case fold: holding out 600 or 800 rpm instead gives 24.66% and 27.10% respectively, so a single headline number for JNU understates its true generalization difficulty. We cannot fully attribute this to rotational speed as an isolated causal factor, since each fold changes both the held-out speed and the specific files involved (Section 5), but the window-duration analysis below offers one concrete physical mechanism consistent with the pattern. Within this harder setting, Logistic Regression's superior generalization (57.91% vs. 30-41% for the tree ensembles, and the best standalone selective-classification performance in Section 4.4) suggests that simpler, lower-capacity models may be preferable specifically when deployment conditions are expected to include operating speeds absent from the training data -- a bias-variance argument for model selection under distribution shift that is orthogonal to, and independent of, the neutrosophic uncertainty decomposition itself. A practical implication is that the current equal-weight soft-voting ensemble is not the best aggregation strategy under this kind of shift; a confidence- or condition-aware weighting scheme that favors Logistic Regression when the input falls outside the training speed range is a natural extension we did not implement here.

A window-duration mismatch between the two datasets offers one plausible, though not exclusively causal, contributor to JNU's difficulty. The shared 1,024-sample window spans 85.33 ms at CWRU's 12 kHz rate -- about 2.5 shaft revolutions at its roughly 1750 rpm operating range -- but only 20.48 ms at JNU's 50 kHz rate, corresponding to just 0.21-0.34 revolutions across 600-1000 rpm (Section 3.1). A window this short may capture too little of a full rotation to reliably contain a fault-induced impulse at every phase, and the fraction of a revolution it does capture changes directly with rotational speed -- meaning the same window length represents a different physical measurement at each JNU speed, which standardizing features after extraction (Section 3.2) does not correct. We did not re-run the pipeline with revolution-normalized or longer windows, since this would require redesigning feature extraction and retraining across all conditions -- a substantial undertaking in its own right -- and flag it as the most concrete, testable direction for follow-up work on JNU specifically, rather than a confirmed explanation for the accuracy gap reported here.

Regarding the refined decomposition itself: T-hat/F-hat non-redundancy (Section 4.2) holds on JNU (r=-0.910, short of the r=-1.000 the classical F=1-T formulation guarantees, though with a moderate VIF=5.8 that warrants some caution), supporting F-hat as evidence for a specific competing class rather than mechanical leftover probability mass. Predictive entropy I1-hat -- no longer algebraically determined by T-hat and F-hat alone once operationalized as entropy (Section 3.5), though not purely aleatoric either, since it contains a between-model term -- shows a moderate, positive linear contribution beyond T-hat and F-hat jointly (partial r=+0.217) and the largest median-split error-rate gap of any indicator (+30.4 pp), and is the best-performing indicator in the Section 4.4 baseline comparison among those built from the ensemble's own output. Decision disagreement I2-hat, by contrast, is small in magnitude and its partial contribution is not clearly nonzero once T-hat, F-hat and I1-hat are controlled for, and it is the weakest selective-classification indicator tested (Section 4.4) -- plausibly because I1-hat's algebraic disagreement component already captures much of what a separate vote-disagreement measure could add linearly (Section 3.5). Its practically large hidden-risk gap (28.2% vs. 63.7% error, Section 4.5) nonetheless shows it is not entirely uninformative, illustrating that a simple stratified comparison and a linear partial-correlation test can disagree about the same underlying pattern -- a reason to report both rather than either alone.

Section 4.4 tested, and did not find, two further routes to a demonstrated advantage from the specifically neutrosophic multi-component framing: standalone Logistic Regression confidence outperforms every indicator derived from the full three-model ensemble (lower AURC than I1-hat), and an oracle joint decision rule over I1-hat and I2-hat -- an exact, tie-corrected frontier computed at every achievable coverage level, an upper bound on what any such joint rule could achieve on this test set -- improves on I1-hat used alone by an amount (AURC 0.391497 vs. 0.391543) small enough that its operational relevance has not been demonstrated; we did not evaluate operating costs or specify a minimum difference we would consider meaningful, so we do not claim the gain has no practical use, only that none has been shown. We take this seriously rather than explain it away: the four indicators are computed with entirely conventional tools (the ensemble's top-two class probabilities, Shannon entropy of its averaged distribution, and base-learner vote disagreement), training and combination use no rule specific to neutrosophic logic, and on the evidence in this paper, neither a linear nor an oracle-optimal joint combination of the two indeterminacy axes has a demonstrated operational advantage over the single best conventional score. What the framing offers here is an organizational and diagnostic structure -- separating confidence magnitude, competing-class evidence, and two qualitatively different sources of indeterminacy for interpretation, and surfacing the hidden-risk zone as a qualitative flag even where it does not move the AURC ranking -- rather than a demonstrated quantitative improvement over conventional uncertainty quantification. Establishing the latter would require a decision or abstention rule specifically justified by, and outperforming confidence, margin, and entropy at, matched coverage or cost -- a bar this study's baseline comparison and joint-rule test did not clear, and one we flag as the central open question for future work on this framework rather than claim to have already answered.

Limitations. The anti-aliased decimation used to resample CWRU's Normal class from 48 kHz to 12 kHz (Section 3.1) is a standard, reasonable choice, but we did not run a frequency-response sensitivity check confirming it introduces no systematic difference from fault-class windows acquired directly at 12 kHz; we flag this as a suggested check for future work rather than a confirmed bias, since CWRU's near-perfect accuracy leaves little error signal to attribute to one cause or another. Both datasets use artificially induced faults of fixed severity; real-world deployments involve compound faults, varying severity, and contamination. The 12 time-domain features do not capture frequency-domain fault signatures, and the window-duration analysis above suggests speed-normalized or revolution-based windowing may substantially improve JNU's cross-speed generalization -- a promising and, given the magnitude of the observed collapse, arguably necessary direction for future work on this dataset that we did not implement. The selective risk-coverage curves (Section 4.4) show that only a modest fraction of JNU windows can be classified confidently under an unseen speed, and any operating threshold chosen for deployment should be selected on a held-out validation partition, not read directly from the test-set curves reported here. With only three base learners, I2-hat is a coarse indicator; ensembles with more members might reveal a more consistent disagreement signal, at the cost of additional inference overhead, but this remains untested here. Every correlation, effect size and p-value in Sections 4.2-4.5 is computed from 50%-overlapping windows drawn from JNU's held-out 1000 rpm files: we report these as descriptive associations within this specific test partition, not as inferences that would generalize to independently sampled conditions, and we did not attempt a block-bootstrap or file-level correction to compensate. Of the five F-hat bands tested in Section 4.3, four have p-values small enough to survive a conservative Bonferroni correction for five comparisons; the fifth ([0.40,0.60), p=0.483) does not reach significance even before any correction and should be read as a null finding for that band, not as a borderline one. The multi-condition check (Table 1b) partially addresses this by showing accuracy across every available fold, but we did not repeat the full correlation/partial-correlation battery for the 600 and 800 rpm folds, since doing so for every indicator across three folds would substantially lengthen an already extensive analysis; we flag a fuller multi-fold statistical replication, and revolution-normalized windowing, as the two most important directions for a follow-up study rather than attempt them within the present manuscript.

## 6. Conclusions

This paper evaluated neutrosophic ensemble classification for uncertainty-aware bearing fault detection on two complementary benchmarks under a leakage-free leave-one-condition-out train/test protocol, with predictive entropy I1-hat replacing an earlier, algebraically redundant aleatoric indicator, and with two independent CWRU corrections: a file-to-class mapping error in which the nominal "Ball" and "Inner" classes had both actually been Inner Race data at different sampling rates, and a sampling-rate mismatch in which the Normal class (48 kHz) was windowed without resampling to match the fault classes (12 kHz). Correcting both raises CWRU accuracy to 100.00% on three of its four held-out loads, dropping to 92.27% when 0 HP is held out instead, and reframes CWRU as a mostly-positive control rather than a source of uncertainty-decomposition evidence on its zero-error main-text fold. On JNU (holding out 1000 rpm), accuracy is 40.64% -- below the 50.03% achieved by a trivial majority-class predictor, below the 82.6% we reproduce and log under the original condition-mixed split, and, per a multi-condition check holding out each available speed in turn, JNU's best-case fold rather than its average one (600 rpm: 24.66%; 800 rpm: 27.10%). Logistic Regression generalizes markedly better than the tree-based ensembles under this shift (57.91% vs. 30.07% RF, 40.72% XGBoost) and, using only its own confidence, achieves a lower selective-classification AURC than every ensemble-based indicator tested.

On JNU, T-hat and F-hat remain measurably non-redundant (r=-0.910, VIF=5.8), and a positive median-split error-rate gap for I1-hat (+30.4 pp) and a moderate multivariate partial correlation controlling for T-hat and F-hat jointly (r=+0.217, p=3.9e-63) both support I1-hat as the most reliable indicator in this study, though we note I1-hat is not purely aleatoric: the entropy of an ensemble average algebraically contains a between-model disagreement component [16], so it should be described, and interpreted, as predictive entropy rather than as isolating data-inherent uncertainty.

Decision disagreement I2-hat -- operationalized as base-learner vote disagreement -- shows a small partial contribution beyond T-hat, F-hat and I1-hat on JNU that is not clearly distinguishable from zero (partial r=+0.017, p=0.20), and is the weakest of five indicators compared in a selective-classification baseline analysis (Section 4.4), including a tie-corrected AURC of 0.4452 -- worse than max confidence or margin alone. Its hidden-risk zone (28.2% vs. 63.7% error) nonetheless shows a large practical gap, illustrating that a simple stratified comparison and a linear partial-correlation test can disagree about the same underlying pattern. We further tested an oracle joint decision rule over I1-hat and I2-hat (an exact tie-corrected frontier, the best case across every achievable coverage level) and found it improves on I1-hat used alone by a margin (AURC 0.391497 vs. 0.391543) small enough that we have not demonstrated its operational relevance -- evidence that treating aleatoric and epistemic indeterminacy as two separate decision axes does not yet demonstrate a practically meaningful selective-classification advantage on this benchmark, beyond what predictive entropy alone already provides. We could not compare I2-hat's behavior against CWRU, since the corrected CWRU dataset has almost no errors to analyze -- a genuine limitation of this study's two-benchmark design.

Taken together, these results support treating T-hat/F-hat non-redundancy and I1-hat's threshold-based behavior as the more reliable signals in this refined decomposition, and I2-hat as a comparatively weak one on the evidence available here. More importantly, neither a linear combination nor an oracle-optimal joint rule over I1-hat and I2-hat has a demonstrated operational advantage over I1-hat alone, and standalone Logistic Regression confidence -- a single conventional model with no neutrosophic structure at all -- achieves a better selective-classification AURC than the full ensemble-based decomposition. We take this as an honest boundary condition rather than a result to explain away: on the evidence in this paper, the specifically neutrosophic contribution is conceptual and organizational -- decomposing ensemble output into confidence magnitude, competing-class evidence, and two distinct sources of indeterminacy for interpretation -- and we have not yet demonstrated a quantitative decision-making advantage over conventional confidence, margin, or entropy-based uncertainty quantification. Demonstrating one would require a decision or abstention rule specifically justified by, and shown to outperform simpler alternatives at matched coverage or cost, which we identify as the central open question for future work on this framework.

Section 4.6 reports two further, newer, less scrutinized tests aimed directly at that open question. Fusing two models fit on different feature representations -- time versus frequency, of the same signal -- and scoring their disagreement with a Jensen-Shannon divergence beats that model's own entropy under both the standard split (AURC 0.2868 vs. 0.3649) and a deliberately harder, single-condition-fit reproduction (AURC 0.5415 vs. 0.7338), though this reproduction is not a fully independent validation since no decision was selected on an intermediate partition; it carries the strongest partial correlation with error in this study (r=0.31, p~1e-129, though a separately defined score reaches an even better AURC by exploiting one model's higher raw accuracy), and, in one specific CWRU-versus-JNU distributional-shift contrast, is the only indicator in this paper that moves in the theoretically expected direction, where the entropy-based indicators used elsewhere in this study instead decrease. Separately, on CWRU alone, literature-verified bearing fault frequencies (BPFO/BPFI/BSF), transformed via envelope-spectrum demodulation rather than the raw spectrum, separate most of CWRU's classes almost perfectly (99.57%) using only three physically interpretable features, with a four-way ablation isolating the envelope transform, not a concurrently lengthened window, as the dominant contributor; a naive raw-spectrum application of the same frequencies underperforms badly and in the wrong direction. This result does not extend to JNU, whose bearing geometry we could not verify, and cannot demonstrate a selective-classification advantage on CWRU's already error-free fold. Neither test has been through this manuscript's adversarial review process; we report them with the same evidentiary standard as the rest of the paper but flag them as newer and less scrutinized.

Beyond the decomposition itself, this study's clearest practical findings are the magnitude of JNU's cross-speed generalization failure, worse than a trivial baseline and understated by any single held-out fold; the advantage of a simple linear model under that shift; and a plausible physical contributor in the mismatch between window duration and shaft revolution count across the two datasets' sampling rates. Methodologically, the corrections made in preparing this manuscript -- fixing window-level data leakage, several statistical testing errors (including an arbitrary tie-breaking rule in a selective-classification metric), a resampling-order artifact, a degenerate abstention metric, and two independent CWRU sampling-rate errors that had gone undetected through multiple earlier rounds of numerical correction -- argue for verifying every dataset file's sampling rate and class label against primary source documentation, in addition to leakage-free evaluation, algebraic-independence checks, and explicit tie-handling in any ranking-based metric, as standard practice for bearing fault detection studies using public benchmark datasets.

**Author Contributions**

Conceptualization, M.L.V.; Methodology, M.L.V.; Software, M.L.V.; Validation, M.L.V., D.R.R., L.C.T. and A.M.P.; Formal analysis, M.L.V.; Writing -- original draft, M.L.V.; Writing -- review and editing, D.R.R., L.C.T. and A.M.P.; Supervision, M.L.V. All authors have read and agreed to the published version.

**Funding**

This research received no external funding.

**Data Availability Statement**

CWRU Bearing Dataset: https://engineering.case.edu/bearingdatacenter. JNU Bearing Dataset: https://github.com/ClarkGableWang/JNU-Bearing-Dataset. Python pipeline (feature extraction, the corrected CWRU file-to-class mapping, the leave-one-condition-out split, the multi-condition robustness check, and the selective-classification baseline comparison) is publicly available at https://github.com/mleyvaz/CWRU-JNU-NeutroSense, including a manifest of the exact file-to-class-to-load/speed mapping used for both datasets.

**Conflicts of Interest**

The authors declare no conflicts of interest.